\documentclass[sigconf]{acmart}

\usepackage{booktabs}
\usepackage{amsmath}
\usepackage{amsfonts}
\AtBeginDocument{%
  \providecommand\BibTeX{{%
    \normalfont B\kern-0.5em{\scshape i\kern-0.25em b}\kern-0.8em\TeX}}}

\setcopyright{acmcopyright}
\copyrightyear{2018}
\acmYear{2018}
\acmDOI{10.1145/1122445.1122456}

\acmConference[DeepNLP '20]{DeepNLP '20}{July 30, 2020}{Xi'an, China}
\acmBooktitle{DeepNLP’20, July 30, 2020, Xi’an, China}
\acmPrice{15.00}
\acmISBN{978-1-4503-XXXX-X/18/06}

\begin{document}

\title{Query Understanding for Natural Language Enterprise Search}

\author[F.\ Borges]{Francisco Borges}
\email{fborges@salesforce.com}
\affiliation{%
  \institution{Salesforce Inc.}
}

\author[G.\ Balikas]{Georgios Balikas}
\email{gbalikas@salesforce.com}
\affiliation{%
  \institution{Salesforce Inc.}
}

\author[M.\ Brette]{Marc Brette}
\email{mbrette@salesforce.com}
\affiliation{%
  \institution{Salesforce Inc.}
}

\author[G.\ Kempf]{Guillaume Kempf}
\email{guillaume.kempf@salesforce.com}
\affiliation{%
  \institution{Salesforce Inc.}
}

\author[A.\ Srikantan]{Arvind Srikantan}
\email{asrikantan@salesforce.com}
\affiliation{%
  \institution{Salesforce Inc.}
}

\author[M.\ Landos]{Matthieu Landos}
\email{matthieu.landos@salesforce.com}
\affiliation{%
  \institution{Salesforce Inc.}
}

\author[D.\ Brazouskaya]{Darya Brazouskaya}
\email{darya.brazouskaya@salesforce.com}
\affiliation{%
  \institution{Salesforce Inc.}
}

\author[Q.\ Shi]{Qianqian Shi}
\email{qianqian.shi@salesforce.com}
\affiliation{%
  \institution{Salesforce Inc.}
}

\begin{abstract}
Natural Language Search (NLS) extends the capabilities of search engines that 
perform keyword search allowing users
to issue queries in a more ``natural'' language. The engine tries to understand the meaning of the queries and to map
the query words to the symbols it supports like Persons, Organizations, Time Expressions etc.. It, then, retrieves the information that 
satisfies the user's need in different  forms like  an answer,  a record or a list of records. 

We present an
NLS system we implemented as part of the Search service of a major CRM
platform. The system is currently in production serving thousands of
customers. Our user studies showed that creating dynamic reports with NLS saved more than 50\% of our user's time compared to achieving the same result with navigational search.   
We describe the architecture of the system, the particularities of
the CRM domain as well as how they have influenced our design decisions. Among
several submodules of the system we detail the role of a Deep Learning Named
Entity Recognizer.
The paper concludes with discussion over the lessons learned
while developing this product.
\end{abstract}


\keywords{Conversational Search, Named Entity Recognition, Enterprise Search, Record Retrieval}
\maketitle

\section{Introduction}

Recent advances in Machine Learning have resulted in considerable improvements
in Natural Language Search (NLS) systems. As a result, several popular web search engines (Google, Bing, Yandex, DuckDuckGo to name a few)
can now resolve queries of various intents ranging from geography (``Which is
the capital of France?'') to restaurant suggestions (``show me sushi
restaurants'') and many others.  Such systems are Natural in the sense that
their style and vocabulary is much richer and more human-like than what one
would observe with keyword search queries. From a user experience point-of-view,
NLS is very powerful: it allows the user to chain several filters together in
natural (colloquial) language, getting them applied in the results. Completing
the same task with keyword search would require issuing a relevant query,  navigating to the retrieved results and
manually inspecting them. In other use cases completing the same task would require manually
creating a report or formally querying a ``database''. In the restaurant example
above, a user can type ``show me open sushi restaurants near me'' and expect the NLS 
results to be filtered according to their opening hours and geolocalization.

In this paper we describe the main components of an NLS system we built as part of the search service for a major Customer Relationship Management (CRM) service. The goal of CRM is to manage a company's interactions with current and potential customers. It accompanies  vendors in every step of the cycle from finding a lead, converting it and later supporting existing contracts. As such, it must be flexible and customizable to be tailored to a variety of customers in very different domains and needs. In addition, data privacy is of utmost importance to a CRM service both at the user level\footnote{The EU's General Data Protection Regulation (GDPR) is an example of law we, and many of our customers,  must comply with.} and at the customer level. For instance, our customers may not want their data to be used for improvements in machine learning systems that will be used by their competitors. Designing an NLS system that copes with these limitations poses several ML challenges in every stage of the development pipeline including training data generation, model selection and online evaluation.

Despite the challenges of the CRM field, providing NLS capabilities in a CRM search service can dramatically improve the workflow of the users. For example, a common workflow of a salesperson is to inspect their assigned opportunities and perform different tasks to advance and convert them. This workflow is accelerated when they can retrieve opportunities in a given state like ``open'' or ``closed''. Same for a salesperson on the road trying to visit her customers: retrieving opportunities given a location can save her time. Our user studies showed that creating dynamic reports using NLS saves more than 50\% of a user's time compared to achieving the same goal with navigational search. Enabling users to achieve their goals faster is the goal of search. We believe that the adoption of our NLS system, that today serves thousands of users, is mainly driven by this acceleration.

Our NLS system consists of several submodules that are responsible for understanding parts of the query, applying business logic, enforcing sharing rules etc.  Among them, there is a deep learning Named Entity Recognition (NER) system that is part of the query understanding submodules. While developing the NLS system we faced different challenges concerning the integration of Deep Learning to our system. Some examples are the lack of training data for our use-case, our need to have a single model serving very different customers and engineering constraints for the size of the model. In this paper we discuss our choices and solutions to these problems.   

The contributions of this paper can be summarized as follows:
\begin{itemize}
\item We describe the core components of an NLS system that is in production and serves thousands of customers daily.

\item We detail the system architecture and the design choices of a deep learning NER
that is among the core systems of our NLS implementation

\item We share our lessons learned and best practices from bringing this service to production and the machine learning decisions we made. We also highlight how these decisions evolved during the lifecycle of our project. We hope that sharing these will benefit members of our community who are building such systems  to overcome  problems similar to those we faced. 
\end{itemize}


\section{Related Work}

We aim to provide our users an interface to interact with their data stored in a database \cite{androutsopoulos1995natural}.
NLS should use  Natural Language utterances instead of a formal language (like SQL) or a controlled language \cite{schwitter2010controlled} like Attempto Controlled English \cite{fuchs2008attempto}. These  require additional knowledge or training and we believe this is an adoption barrier. 

To develop the NLS functionality in our search service, we position ourselves in the field on semantic parsing for Natural Language Search \cite{kamath2018survey}. 
Semantic parsers have found applications in various domains from robot instructions following \cite{artzi2013weakly,williams2018learning} to conversational search and dialog management \cite{raghuvanshi2018developing,srivastava2017parsing}
Our implementation borrows ideas from \cite{liang2015bringing,liang2016learning}. 
Internally, NLS uses a grammar in order to generate training examples for our Named Entity Recognizer and a more semantically and syntactically narrow version of this grammar to support autocomplete and query suggestions.  
The role of the grammar is different in our system compared to these papers. We do not validate the logical forms with the grammar. We have a system in place that validates it by applying our business logic which controls for several aspects of the generated logical form including privacy. 
More recent work has shown promising results  using end-to-end neural networks  for semantic parsing \cite{dong2018coarse,zhongSeq2SQL2017}. Instead of relying exclusively on a neural network, we have designed our system with several sub-components to allow for complete explainability of our results.

In the context of semantic parsing, previous works have used grammars in order to extend the training set. In \cite{wang2015building} Wang et al. describe their approach where they begin by  manually  generating representative logical forms paired
with canonical utterances for a given domain. They, then, use crowdsourcing to extend the produced data points by paraphrasing. In our domain we do not massively label data,  such as production logs, due to privacy and legal issues. Another similar approach is recombination \cite{jia2016data} where a synchronous Context Free Grammar is learned by  identifying conditional independece assumptions in the training data. This grammar allows to sample more training data thus extending the initial training set. Recombination is similar to our approach in that it allows sampling  training instances from a grammar by extending an existing,  current dataset. In our case, however, we did not have any initial training data: we used a probabilistic Context Free Grammar to bootstrap the initial training set. 
 
Developing NLS we came to understand that such a deployed system should come with autocomplete and query suggestions support. We found that, instead of user manuals, query suggestions are much more efficient to educate the users for the capabilities of our search service. The value of query suggestions has been found by previous studies e.g., \cite{jagadish2007making}. Query completion and suggestion systems have also been proposed for formal languages such as SQL \cite{khoussainova2010snipsuggest} or SPARQL \cite{bast2017qlever}.
Query suggestions as part of a semantic parser is also discussed in \cite{arkoudas_semantically_driven}. 
The most important difference is, perhaps, that our customer data are kept apart and siloed for each customer; scaling approaches as those described in  \cite{arkoudas_semantically_driven} to treat each customer separately to thousands of them is an important engineering challenge in itself. 
Instead, until now we have opted for using   a probabilistic  grammar for suggestions, which only needs probability weights that can be controlled for each organization by the metadata. 

\section{Problem Statement}
\begin{figure*}[t]
\includegraphics[scale=0.4]{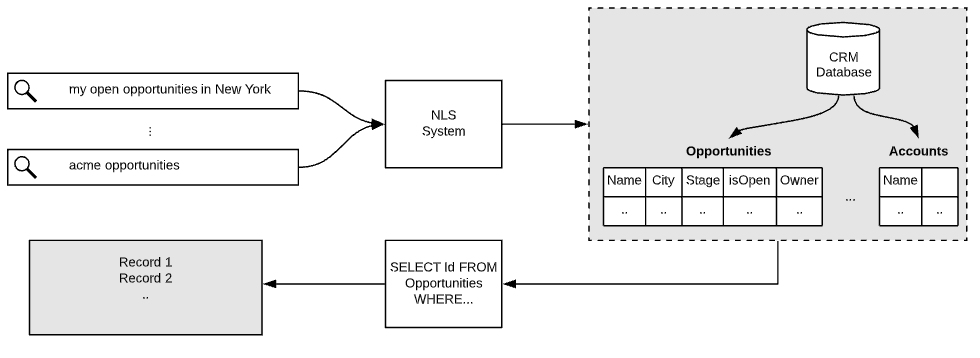}
\caption{Examples of processing NLS queries. The NLS system should understand that the issued queries are not keyword searches and construct SQL-like statements. These statements  when executed against the CRM schema return lists of records. In the examples NLS should return lists of opportunities that satisfy the different filters that are described in the user query. For simplicity, we do not show the annotations that NLS should return to the UI to explain the query interpretation. }
\label{fig:toyUseCase}
\end{figure*}

The core goal of our NLS system is to take short, colloquial and often grammatically incorrect  queries and convert them into formal structured ``database'' queries. This allows users to filter and select information through the CRM schema that organizes their data without using a dedicated formal language.

We position ourselves in an Information Retrieval setting where the user expresses their need as a query. We treat each query independently leaving the concept of search sessions out of the scope of this work. The NLS system we describe extends our deployed search engine by distinguishing between two different user intents. The first user intent is keyword search. The second is NLS. Figure \ref{fig:toyUseCase} illustrates examples of NLS queries that our system is expected to resolve.

The user of Fig. \ref{fig:toyUseCase} issues the first query (``my open opportunities in New York") with a goal to fetch a filtered set of the records stored within their CRM system. The query differs from a keyword search in that it expresses an intent to select a subset of the ``Opportunities'' records,\footnote{In the rest of the paper we refer to CRM entities, like Opportunities and Accounts, as CRM objects or CRM entities interchangeably.} after applying the prescribed filters: open refers to the boolean ``isOpen" and ``New York" can refer to a City or a State. The system must recognize that the query is to be resolved using the ``Opportunities'' entity (our CRM schema has several entities) and then construct an appropriate query that will return the expected results. The second query (``acme opportunities") needs to be resolved differently: the system should recognize that ``acme" is an organization, find its record id in ``Accounts" and return the opportunities whose account id matches the resolved id for ``acme". For each user query an appropriate database query needs be constructed to return to the user a list of records of the Opportunity entity.

We frame NLS as a semantic parsing problem \cite{kamath2018survey,liang2015bringing,liang2016learning}: given a natural language utterance (the query) we try to generate a semantic representation. Theoretically the query can be of arbitrary length combining several concepts like a record type or a time expression. The final semantic representation is a logical form that (i) allows querying the objects of our CRM schema via an SQL-like language\footnote{In this paper we refer to both SQL-like and formal languages interchangeably} and, (ii) provides enough annotations for the UI to explain the retrieved results to the users.

In the remainder of this section we describe the challenges and the constraints of CRM in more detail as they have influenced several of our design decisions and machine learning choices.

\subsection{Challenges of enterprise CRM search}

We provide enterprise search services to more than 150 thousand CRM customers. Our customers\footnote{We refer to a customer also as an organization or a CRM customer.} range from small and medium sized businesses with few people, nonprofits with 100s of volunteers to large companies with 100s of thousands of employees. Each customer has a number of different users and most importantly its own private data. In this section we will describe some of the challenges that arise from the CRM domain and, more generally, from enterprise search \cite{hawking2004challenges,dmitriev2010enterprise}. 

\noindent\textbf{Privacy} Our enterprise search service is characterized by the fact that customers CRM data is completely siloed from one another. Data is not only indexed separately but statistical aggregation is limited within each silo. Privacy is manifested both in the user and content level and is extremely strict. We do not  customer queries to train models, nor do we inspect customer CRM  data to understand its characteristics better. Many customers are unwilling to give us contractual permission to blindingly mix customer CRM data, such as names of Accounts or just data mined from query logs to build models. 

\noindent\textbf{Usability} Our NLS offering augments our search services with the capacity to interpret NLS queries. These queries are entered into the same search input box as regular search. Reusing the input field has many practical consequences. The first is that users may not be aware of the new NLS capabilities. The system must offer NLS query suggestions and NLS query auto-complete features to help them. When a query is interpreted as NLS, the User Interface (UI) must provide clear explanations of how the query was interpreted, as well as providing fall-back and remediation mechanisms. A fall-back is when the user demands a query to be interpreted as a keyword search. Remediation is when the user needs to change the interpretation of a query. For example, when our system resolves a Person in a query to a given record id the user may wish to correct that by selecting another Person record. Recall the example of Fig. \ref{fig:toyUseCase}: NLS needs to resolve ``acme" to an Account record id but there can be multiple records with ``acme" in their name. The user should be able to signal that ``this is not the acme I meant'' and be able to select another record from Accounts. The requirement to explain the query's interpretation and to be able to change specific clauses in it must be taken into account when designing a solution. For example, an end-to-end deep learning model with limited explainability is not a suitable solution.

\noindent\textbf{Scale} One of our key requirements is to deliver NLS at scale, to each of our customers. This requires a generic architecture and generic models. We are not building machine learning \textit{a la carte} for each customer but a system that delivers NLS to all of them at once. Yet this system must be able to deal with each customer's data, CRM workflows and even their customized CRM schemas. For instance, many of our customers manage record ownership differently: ``my accounts'' can have a different meaning depending on the customer and our engine must be able to interpret the query according to the customer's specific record management workflow. An option is to require each customer's administrator to configure the system at length. We believe that this is a huge barrier to adoption. Therefore, another key requirement is that our NLS has to work and deliver customer value as is, without requiring admins to invest time and energy pre-tuning their system.


\section{System Overview}\label{sys_overview}

\begin{figure}
\includegraphics[scale=0.35]{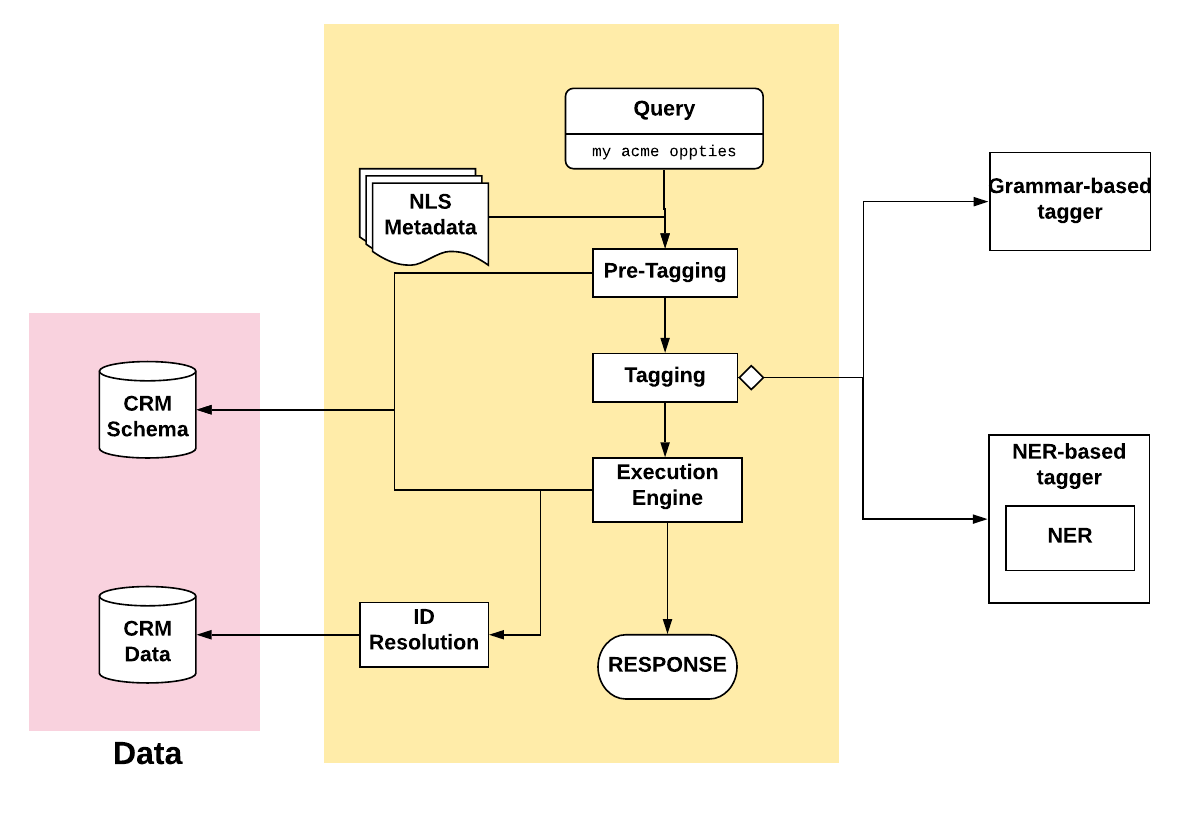}
\caption{A simplified representation of our NLS system showing the code organization into submodules and where each comes in during the handling of a query.}
\label{fig:system}
\end{figure}

Our natural language search service implements and uses two different APIs: (i) query processing and (ii) query suggestion. Query processing is responsible for processing a user query and trying to transform it into a formal structured query. It  returns results and a data structure that explains how the query was interpreted. Query suggestion generates valid NLS query autocompletion and suggestions to the user. In this overview we  describe our query processing system in detail.

The query processing system is organized into several submodules roughly as shown in Fig. \ref{fig:system}. There are two groups of submodules in our system: the first group is responsible for ``understanding'' the query and the second is responsible for processing and validating the results of the first. Query understanding steps involve pre-tagging, tagging (named-entity-recognition) and parsing their results into a semantic tree. Processing and validating steps involve sanity and security checks, rules enforcement, ID resolution and other steps necessary to generate results and explain what was done to our users.

When a query arrives, it is associated with the NLS metadata. These two elements are shared over the entire handling of a query. The metadata carries all configuration relative to the entire system and ensures that the whole system behaves consistently. Consistent behavior means that when we change the configuration of a certain aspect of our system the change is reflected consistently across all different submodules.

The query first goes through Pre-Tagging where multiple expert systems are applied. An expert system is a component that handles a specific sub-domain such as identifying references to ``picklist'' CRM fields and its values.
The query is then passed on to a tagger which attempts to parse the query into one or several semantic trees.
The validity of these trees is evaluated by the Execution Engine. This involves requirements and consistency checks, ID resolution and finally applying security rules against each concept.

We built two taggers: one based on a simple grammar, called the Suggestions Grammar, and another based on a Deep Learning NER. Both share most of the steps described above.
The system first attempts to interpret the query with the Suggestions Grammar. If no semantic tree is valid, it runs the DL NER tagger. The DL NER system can interpret a larger variety of queries than the Suggestions Grammar but it can be unpredictable at times. It can also produce multiple trees that the tagger orders by priority.

In the following subsections, we describe each of the subsystems in greater detail.

\subsection{NLS Metadata}

The NLS Metadata is an object that unifies access to all configurations of our system. It ensures consistency across the submodules and allows developers to easily inspect the system configurations at once. Another practical aspect of a dedicated metadata system is to coordinate and implement A/B experiments. When a query is to be handled as part of an A/B experiment, we just pass the corresponding experiment metadata.

\subsection{Pre-tagging}

The addition of pre-taggers to our system happened relatively late in its development. The pre-taggers are expert systems that handle recognition tasks that require customer-specific data. Pre-taggers are key components in making NLS work well for thousands of customers with siloed data and different vocabularies with just a single Deep Learning NER model.

\begin{figure}
\includegraphics[scale=0.33]{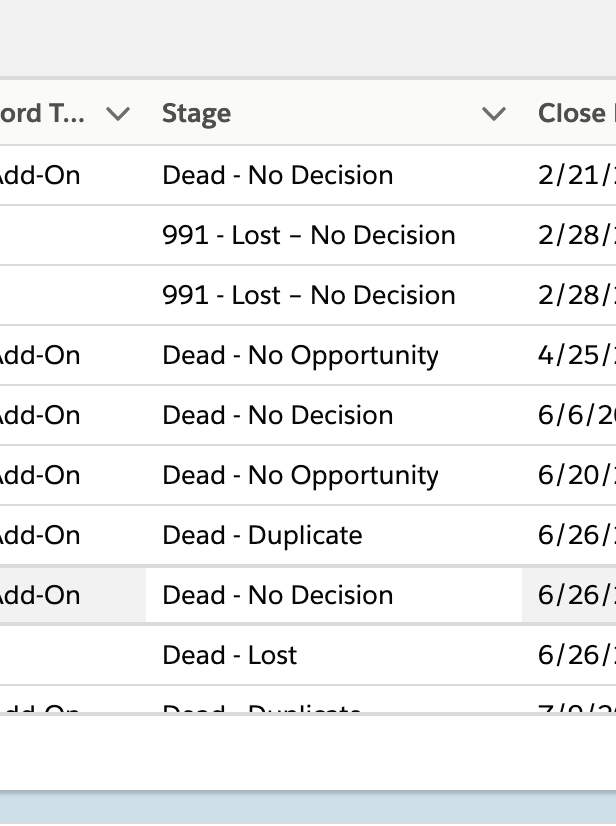}
\caption{An illustration of how recognizing picklist values can be difficult in real life. Lots of values are very similar. The values shown are from our own CRM organization and not from any of our customers.}
\label{fig:oppStage}
\end{figure}

One of our pre-taggers recognizes picklist values related to specific entities.
Picklists are dropdown fields that  assume values from a list defined in the CRM schema. While there are predefined values for them, each organization can easily add, remove or edit/rename them to serve their purposes. An example of a picklist that a Deep Learning solution can handle well is a list of countries. Another example is the stage of an opportunity as Figure \ref{fig:oppStage} shows. In the second example, a DL system cannot easily generalize to recognize such unseen values as they are specific to an organization and normally extremely different across organizations.

Another pre-tagger in place is responsible for recognizing references to entity names. Our CRM platform allows customers to rename our standard CRM entities and many customers use that feature. In the example of Fig. \ref{fig:oppStage} for example, a customer could have renamed ``Account'' to ``Customer'', ``Factory'' or, say, their translation to another language. NLS should be able to correctly resolve queries even with renamed entity names. While our deep learning NER generalizes poorly in such cases, an expert system (in this case it can be a dictionary look-up)  with access to the CRM scheme during query resolution can support the use-case.

\subsection{Tagging}

Our taggers are responsible for recognizing named-entities in a query that are then parsed into semantic trees. As described above, we have two different tagger implementations, the Suggestions Grammar, and the NER tagger. We continue by describing the Suggestions Grammar and the need to have it in place in addition to the NER tagger.

\subsubsection{Suggestions Grammar tagger}
As part of our overall user experience, the query suggestion engine helps users craft queries and drive adoption of NLS. We do not address in detail the Query Suggestion (QS) engine design in this paper, but we need to stress out the interaction with the NLS system. QS must be consistent with the NLS capabilities: its implementation should avoid logic duplication and a suggested query must be interpretable and executable by the NLS execution engine.

Our QS engine is based on a Probabilistic Context Free Grammar (PCFG), the Suggestions Grammar. The system uses that PCFG to generate a Directed Acyclic Graph (DAG) in memory, at query time. This DAG is used to both parse and complete the input query. This approach allows us to deal with organization and user specific vocabulary. We have a single global PCFG described in metadata, and user specific DAGs generated using the query context (user, organization, vocabulary etc). This not only enables us to respect customer privacy, but also provides personalised vocabulary and ranking in query suggestions, while only maintaining a single grammar.

To avoid duplicating logic, the QS engine re-uses part of the NLS pipeline and is controlled by the same metadata driving the NLS engine. The PCFG re-uses NLS pre-taggers logic when possible. Also, the suggested queries go through the same security checks to eliminate candidates that would not be accessible to the user.

When a user selects a suggestion, the query is interpreted by the NLS query processing pipeline. Query suggestions are built to be interpretable, but sometimes deep learning can be unpredictable and fails at tagging suggested queries. Our solution was to turn the suggestion's DAG parser into a second tagger. This tagger has the capability to generate a semantic tree for each generated suggestion. This ensures consistency between generation and interpretation. While it may seem redundant to retag a suggested query, this design choice was critical to consistently interpret both suggested and typed NLS queries.

The Suggestion Grammar tagger is robust but limited in the variety of queries it supports. It generates a single parse tree with no room for misinterpretation.
The NER tagger, described next, supports the whole range of NLS queries.

\subsubsection{Deep Learning NER tagger}

Our main tagger is based on a DL NER model. This model recognizes and tags a wide range of NLS query forms.
The NER model encapsulates a lot of the intelligence in the algorithm. It selects the most probable sequence of tags for the query terms.
The design of the NER model is addressed in details in Section \ref{section:ner} below.

However, the tagger as a system, is more than the NER model.
It generates multiple candidates based on the tags from pre-taggers, and selects the best candidates. Currently, the ordering criteria is very simple, assigning a higher score for the longest matched sequences.
Calling the NER model involves several steps as follows:  
\begin{itemize}
\item The tagger prepares the input query for the NER model, 
\item Pre-tagged terms are masked with a fixed token that the NER is aware of,
\item Some tokens are normalized: numbers and IDs are replaced with corresponding fixed tokens.
\end{itemize}
The NER model output is a sequence of tags for each term.
The tagger filters out the invalid sequences and transforms each sequence of tags into a semantic tree. Invalid sequences include sequences breaking structural rules we manually defined.
Each remaining tree corresponds to a different interpretation of the query. The tagger ranks them in priority order and the topmost interpretations are returned to the Execution engine. Here again, the order is based on manual structural rules.

\subsubsection{Semantic trees}

The output format of our taggers started as a bag of named-entities and evolved into a semantic tree at the time that we added support to entities that modified the meaning of other entities than the root of the tree.

\begin{figure}
\includegraphics[scale=0.48]{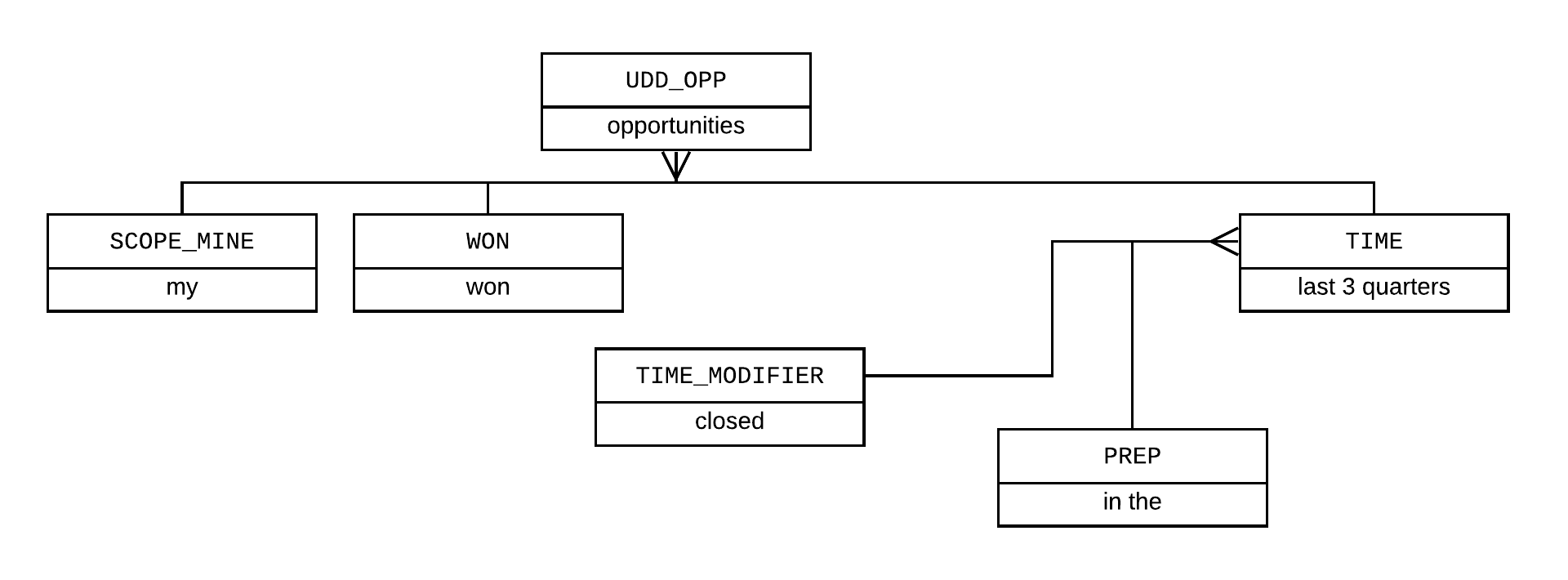}
\caption{A very simple illustration of a semantic tree build for the query ``my won opportunities closed in the last 3 quarters".}
\label{fig:tree}
\end{figure}

Figure \ref{fig:tree} shows an example of a semantic tree for the query ``my won opportunities closed in the last 3 quarters''. One of the challenges of the example is in handling  ``closed''. It can either refer to an ``opportunity'' or it can be attached in the time expression. Creating a tree and attaching it to the time expression disambiguates the semantics: the user is asking for opportunities whose close date is within the last 3 quarters.

\subsection{Execution Engine}\label{sec:executionEngine}

Once we have a semantic tree of representation of a query, it gets passed to our Execution Engine. This engine determines whether the semantic tree is a valid natural language query. It then creates a logical form with enough information to (i) create a structured query based on the customer's CRM schema and (ii) explain to the user how the query was interpreted.

Determining whether the semantic tree is valid involves validating the internal consistency of the semantic tree, not just generally but according to the customer's CRM schema and to the security access rights of the user running the query. If there are any Person or Organization named entities in the semantic tree, they have to be resolved to a record id. Any unresolved references invalidate the tree.

This final secondary structure must also contain enough information for our UI to modify the interpretation of the query should the user decide to do so. Suppose the user issues a query for ``acme's opportunities'' with ``acme'' being tagged as an Organization. If our ID resolution module resolves it to the wrong Account ID (consider that there may be hundreds of valid Account records with ``acme'' in their title), our users can use the remediation mechanisms NLS provides  to select a different Account record.

\subsection{Security}

In our CRM product customers can restrict user access not just to specific records but also to entities and to specific fields of a given entity. That means that some users may not be allowed to view or know about the Account entity and that other users may know about Accounts but not about certain fields of it, like Account.Price.

So both when creating query suggestions as well as interpreting queries we need to enforce these access rules. Which means that besides usual record access checks, we also perform security checks at the concept level.

If a user is not allowed to see the Account entity, our system will not show any suggestions based on this entity nor interpret any queries based on suggestions. For a user that can see the Account  entity but can not see a particular field of it, say Account.Stage, then only the queries that involve this concept will be filtered out from their suggestions and the interpretation of their queries.

\subsection{ID Resolution}

When a query makes a reference to a Person or Organization, e.g. "jane's accounts" or "acme closing opportunities", we must resolve the Person or Organization reference to a record id.

Id resolution is a navigational search problem \cite{broder2002taxonomy} where there may be hundreds of matches to the reference but only 1 of these is valid. In the example of ``jane's accounts'', we may have hundreds of contacts or users named Jane but only one of these records satisfy the user's need. This is different from the classical exploratory search situation where there is normally a continuous range of relevant results.

To complicate things further,  the expected  record  for ``acme'' or ``jane'' will differ depending on who is issuing the query. For example, imagine a sales agent working with an account named ``Acme Canada Ltd'' and another sales agent working with an account named ``Acme Netherlands BV''. The correct record to return for a reference like ``acme'' depends on who is issuing the query.

Resolution takes place via cascade of different machine learning models, some deep-learning based but not all. Due to its complexity and the theme of this paper, we do not elaborate further on it.

\begin{figure}
\includegraphics[scale=0.70]{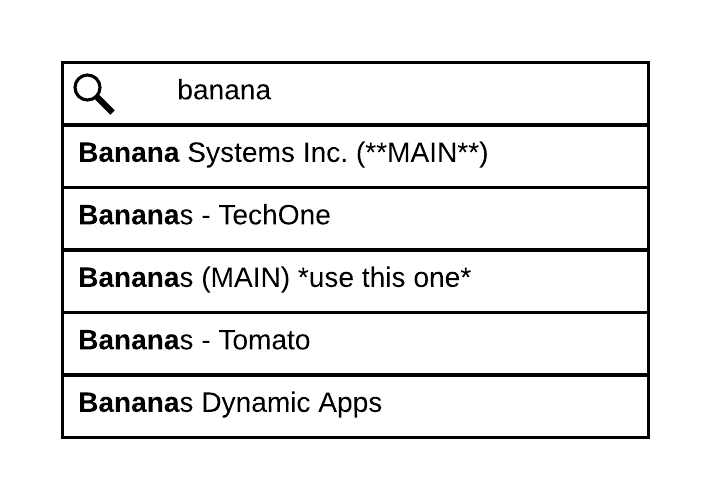}
\caption{An illustration of the typical ambiguity found when resolving references to IDs within large organizations. The user is likely only interested in a single record, there is no gradual continuous relevance in this context. The correct result also depends on which is the user issuing the query.}
\label{fig:ambiguity}
\end{figure}


\section{NER with Deep Learning}\label{section:ner}

Named Entity Recognition is central to any Language Understanding system. For our case, we train a deep learning NER system to distinguish between standard entities like Persons, Organizations, Time and Location (we split location in City, State and Country) some entities that are boolean filters specific in our CRM data model, and the standard CRM schemas that our CRM system uses and we support. This entails that currently we only support NLS queries that can be resolved using the subset of the CRM entities we support and our system does not generalize to arbitrary entities \cite{zhongSeq2SQL2017,lyu2020hybrid,guo2019content}.

\subsection{Data Generation}\label{sec:dataGeneration}
\begin{figure}\centering
\includegraphics[scale=0.53]{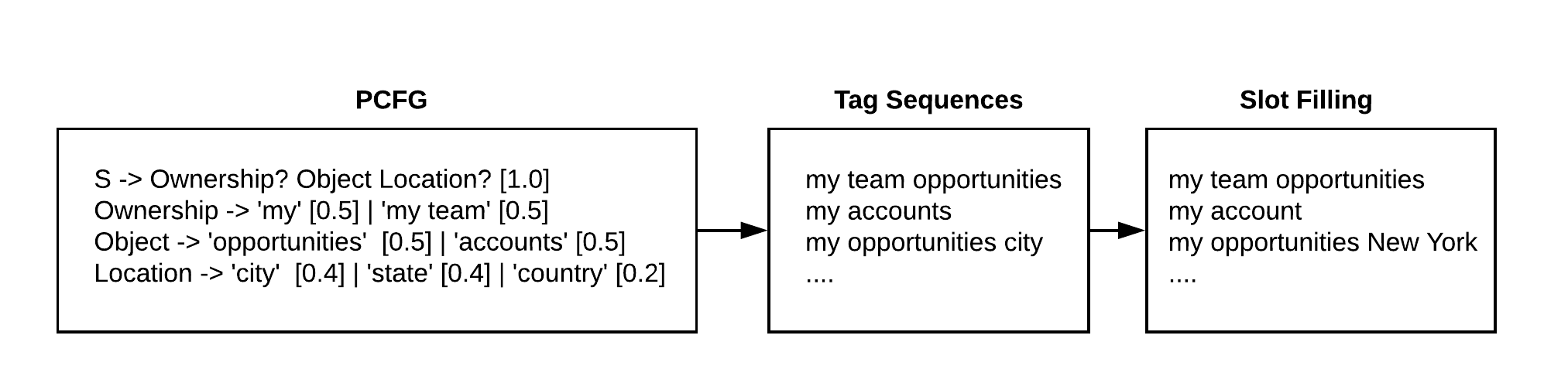}
\caption{Pipeline to create synthetic data.}
\label{fig:syntheticData}
\end{figure}

Before describing the architecture of our NER model, we describe the main problem we faced: how to train a NER model without training data.
In the early development stages of the NLS system, we faced the challenge of collecting data to train the model. Publicly available pre-trained models or datasets are not sufficient for our use-case because:

\begin{itemize}
\item The entities we are interested are a superset of those that are often used in the literature
\item Most available datasets are very different in terms of style. We want to recognize entities in short queries without long-range dependencies and formal syntax.
\end{itemize}

To overcome this problem we decided to use a grammar. Similarly to the Suggestions Grammar, we developed a probabilistic context free grammar which models the order and the concepts of an NLS query. This is done offline, using the Natural Language Toolkit \cite{nltk_2009}.  Figure \ref{fig:syntheticData} shows a simple example of a toy PCFG grammar and a few of the related productions. Notice in the figure that the grammar produces tag sequences that are  populated with appropriate vocabulary.

Having access to synthetic yet realistic tag sequences, we use in-house available collections of organization names, locations etc to populate the tag sequences. Such a slot-filling approach enables us to massively produce training data for our NER model. The data are tailored to our product needs and resemble sufficiently the queries we expect our users will issue.

\subsection{Model Architecture}
\begin{figure}
\centering
\includegraphics[scale=0.33]{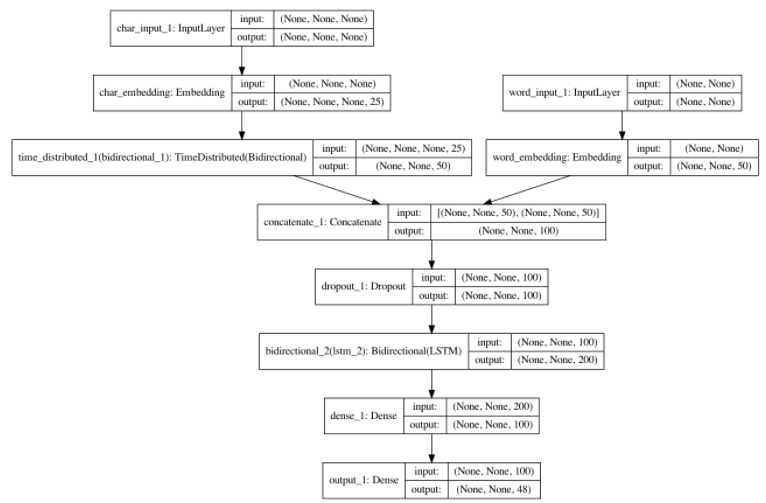}
\caption{The  architecture we are using for our deep learning tagger that is implemented using Keras. We use word embeddings initialized with GloVe and also character embeddings randomly initialized.  }
\label{fig:ner_arch}
\end{figure}

Our NER is a sequence to sequence  Bidirectional-LSTM network that uses word and character embeddings \cite{lample2016neural}. The goal is to predict the tag of each word. Our data use the IOB format (In-Outside-Begin) to denote multi-word entities.
 Figure \ref{fig:ner_arch} shows the architecture as implemented in Keras \cite{chollet2015keras}. We have relied on the open-source implementation of Anago \cite{anago} that we extended for our usage.

 We expect our NER model to fully understand the words in a query. Given their tags, the execution engine generates the logical form that can be used to retrieve the results for the user as described in Section \ref{sec:executionEngine}. Depending on the tags, terms are treated differently. For example, Persons or Organizations are resolved to an id. Due to this strong dependency between the NER tags and the system's response, we require every term of the query to be correctly tagged. For this reason, we decided to measure and optimize our recognition performance against the Strict Correct Ratio (SCR) on the query level and not the $F_1$ score that is often used and is calculated at the tag level. Given an input query with tags $y = (y_1, \ldots, y_n) $ and predicted tags $\hat{y} = (\hat{y_1}, \ldots, \hat{y_n})$  the SCR is 1 if every tag is predicted correctly: $\forall i:  (y_i == \hat{y_i})$. We implemented monitoring callbacks and early stopping using SCR in order to optimize for it.

As seen in the model's output layer in Fig. \ref{fig:ner_arch} there are 48 output classes (24 named entities). Compared to well studied datasets like the CoNLL NER dataset \cite{sang2003introduction} or the WNUT datasets \cite{derczynski2017results} we have many more tags.  This is due to our decision to maintain the complexity to the deep learning system instead of using several subsystems. As a consequence, the deep learning system needs to  distinguish between standard entities (Organization, Person, etc.)  and domain-specific ones like boolean  filters that can be applied when fetching records belonging to specific CRM entities. Recall the example of Fig. \ref{fig:toyUseCase} : ``open" is a boolean filter that must be recognized in the context of ``opportunities". The same term can have a different or no meaning in the context of another object, like ``account" for instance. Modeling such contextual information is a key challenge of our use-case.

\subsection{NER Model Evaluation}

Data preprocessing is central to any machine learning application. Our preprocessing decisions are evaluated  on different dimensions. Two of the most notable ones are NER performance improvement and memory when the model is serialized. The role of the first is clear: better recognition rates ensure better user experience. The second is enforced  by our engineering platform constraints. The NER model is trained with Tensorflow \cite{abadi2016tensorflow}. To serve the model we need it to be less than 200MB serialized. Respecting such a hard limit introduces different trade-offs between model expressiveness and memory footprint. To this end, we have evaluated different aspects of the model complexity: (i) the dimensions of our word and character embeddings, (ii) the size of our vocabulary, and (iii) the depth of the Neural Network too. In the remainder of this section we demonstrate the impact of different hyper-parameters for the architecture presented in the previous paragraph.

\begin{figure}
\centering
\includegraphics[scale=0.25]{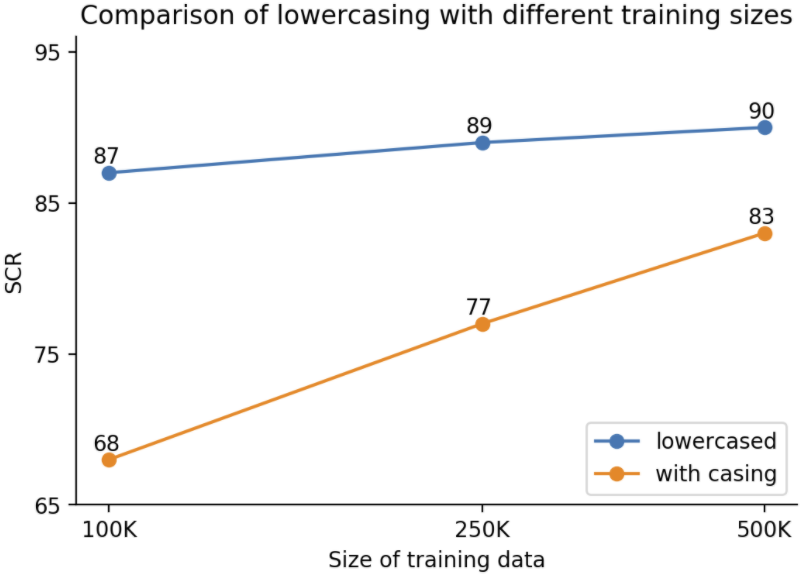}
\caption{The effect of increasing the size of the training data in the performance of the model. Lowercasing ignores casing information by lowercasing the queries. ``With casing" uses separate character embeddings depending on the casing in the query. }
\label{fig:style}
\end{figure}

\textbf{Training Data} For machine learning systems and in particular for deep-learning the size of the training data is an important factor. Using the method of Subsection \ref{sec:dataGeneration} we can practically generate large amounts of training data. Figure \ref{fig:style} shows the performance of NER in terms of SCR with respect to the size of the training data. Increasing the training data improves the performance of NER. As a result, having a high quality PCFG to generate realistic and complex data, can benefit the performance online.

\begin{figure}
\centering
\includegraphics[scale=0.3]{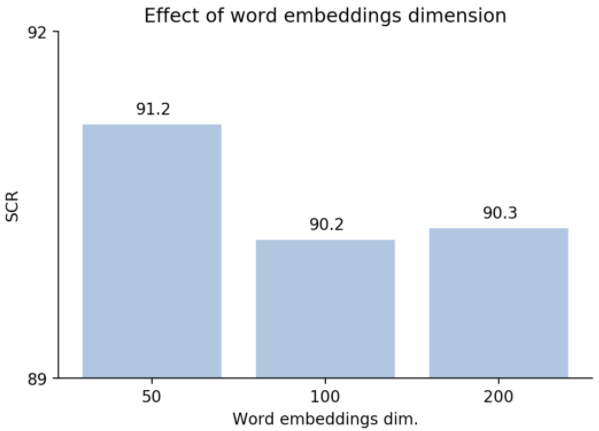}
\caption{Impact of the dimension of the word embeddings in the performance of the NER. For NLS queries, we achieve the optimal performance with embeddings of size 50.}
\label{fig:effectOfEmbeddings}
\end{figure}

\textbf{Embeddings} We initialize our embeddings using GloVe \cite{pennington2014glove}. However,  our vocabulary is different from Wikipedia as it contains a lot of person and organization names. As a result, we observed a lot of out-of-vocabulary (OOV) words. During training we initialize OOV with embeddings randomly from a uniform distribution in [-0.5, 0.5]. Further, we train the embeddings to fine-tune them for our application. Figure \ref{fig:effectOfEmbeddings} shows the effect of the dimension of embeddings. For our application, we find that the best performance is obtained with dimension 50. We hypothesize that this is due to the nature of the task: queries as short text spans, they do not follow grammar rules and there are few long dependencies. Small embeddings can encode the information needed for the task. This finding is convenient for our use-case as it reduces the memory requirements of the deep learning model.

\textbf{Text style} The seminal work of \cite{sutton2012introduction} shows that style information is important for NER and the importance of capturing text style in the form of capitalization.
In the early stages of our development, we evaluated this finding for our data. We implemented casing features as separate character embeddings for the same character. In this experiment, we kept the word embedding lookups lower-cased to respect the memory requirements.
As Figure \ref{fig:style} shows, we found that lowercasing text did not degrade the NER performance.

An interesting observation concerns the impact on the user experience of capitalization features. As expected from a modeling perspective, we found cases where changing capitalization resulted in different NER tags due to capitalization signals in the training data. Such behaviors, however, in an online system can be frustrating for the users as they would receive different results depending on how they type the query. Given this observation and the performance benefits of dropping text style information, we are lower-casing the queries.

\section{System Evaluation}

Releasing a new version of our NER impacts the user experience as it is used to recognize the main parts of the query. Before releasing a model, we follow a thorough performance validation process on several test datasets. The goal is to avoid introducing regressions that may impact negatively the user's trust on NLS and track the quality improvement.

Evaluating the performance of a model in the setting of a product that constantly evolves is a challenge in itself. We cannot use a static validation or test set. Also, as the system becomes more mature, measures need to be taken to ensure smooth and consistent user experience. As a result, at the very early stages of development, one could allow faster model experimentation. However, as the user base increases and starts to rely on our functionality the validation around every change must become stricter.  Apart from the usual validation and test sets that are automatically generated as described in the previous section, we have developed three additional mechanisms to ensure quality control. We present them ordered, from those we implemented early to those we implemented as the product became more popular.

\subsection{Non Regression Dataset}

Our Non-Regression Dataset (NRD) is a dataset that consists of queries whose tags must be predicted correctly always. The concept of NRD extends the concepts of unit tests and test datasets for training a model to serve customers. It is a collection of test cases gathered in a dataset (like a test dataset) but if any of them is tagged incorrectly the training process fails. Having NRD in place ensures that NER tags a set of queries we deem very important as expected. For example, we include generic queries listed in the documentation that lots of users are bound to try.

\subsection{Golden Standard Dataset}

While the NRD ensures that a set of queries consistently succeed, it does not evaluate the quality of the predictions.  To achieve this, we are maintaining a manually curated test dataset, which we refer to as our Golden Standard Dataset (GSD). This is a versioned collection of queries of two categories: queries that should trigger an NLS and those which should trigger a keyword search.

NLS queries are annotated with (i) their NER tags, (ii) their SQL-like statement we expect that NLS will generate and whose results will be rendered to the user and, (iii) one, or several query annotations - which indicate various semantic information like the origin of the query, or the set of concepts in the query. 

The NER tags allow us to evaluate the quality of the NER predictions. NER, however, is a sub-component of the system. The SQL-like statements let us evaluate the whole NLS system as it controls the content that is to be rendered in the user.

Adding query annotations  to each query solves the problem of comparability of systems that are trained on evolving training and dev datasets. The training sets evolve to encompass new features. As the number of tags to  the training data changes to cover more use-cases, predictions may change. Having query level annotations  allows for historical comparisons of model quality: we can always report the performance on queries annotated as ``core-concept'' and compare whether newer models improve on these queries. For example, we are interested to know that the model versioned 0.1.5 achieved 90\% SCR in the queries annotated as ``time-modifiers'' and that this is an  improvement by 6 absolute points compared to the 0.1.4 model that scored 84\%. This insight enables us to ship the model 0.1.5 in production.
At the same time, with each new feature we support, we add queries in the GSD to monitor the system's performance on them.
Finally, such a system allows us to track improvement on regressions or bugs: for every acknowledged case, we can create feature tags pertaining to a bug and ensure that future models improve on it.

\subsection{Model Deployment}

The NER tags can be used to test a new version of a NER model without any deployment. However, to run the GSD evaluation, we need an environment with realistic data.
The GSD contains queries that are based on internal production data. To calculate measures on the SQL-like level, we set up an internal sandbox instance of our internal CRM --our use of our own product-- that is populated with real production data. In this sandbox we issue the GSD queries, collect the logs from the different steps of the NLS system and report on metrics on the entire NLS system.

Evaluating new and existing models in this end-to-end way let us define a release cycle. Before shipping a new model in production, it goes through the different quality gates and it has to perform better or the same in the end-to-end evaluation. In this way, we can also deploy multiple models to be evaluated with A/B experiments.

\section{Lessons Learned}

Our current System is the result of effort across many iterations. During this journey, we learned a lot and our system evolved in some unexpected ways.

\noindent\textbf{Query Suggestions}
In this paper, we only briefly talked about query suggestions. However, they are essential for feature discovery. We observed that concepts that were not exposed in query suggestions were barely used by the users as they were not familiar with them. Even today, with increased adoption, the majority of NLS queries come from suggestions. A core conclusion is that any feature supported by our NLS query processing engine must be supported by query suggestions. Another aspect is that suggestions ranking needs to maintain a critical but delicate balance between user education/new feature advertisement and click-through-rate (essentially a diversity/quality problem on its own).

\noindent\textbf{Deep Learning Model}
Deep-learning is not the ideal tool for every problem, but it is strong enough to handle all tasks well. In our early development process we started with a single tagger that used the deep-learning NER model. This allowed us to bootstrap a working NLS system end to end in a short amount of time and also allowed us to pilot the system early.

As we monitored the system usage and the system failures in particular, we identified concepts that are better served by expert systems, which in the majority of cases are rule-based annotators. While maintaining the core NER component, we added these expert systems to increase robustness and coverage.

Another important reason in favour of the expert systems was the variety of data across customers due to their ability to extensively customize our CRM platform. The demand to support more use-cases, in diverse settings, with restricted access to customer data for training makes it challenging for the NER to generalize well. On the other hand, expert systems that have access online to the configuration of an organization are more robust for simple recognition tasks (like boolean filters or picklist values).

Additionally, in our setting, the DL-based algorithm had to be complemented by a simpler and more robust grammar based algorithm - the Query Suggestions Grammar. This originated from the need to consistently support suggested queries execution but it finally provides a good balance between the two systems. The Query Suggestions Grammar  alleviates pressure on DL because it handles easy queries predictably and without surprises. DL alleviates pressure from Query Suggestion Grammar-based algorithm because it handles hard complex queries and idiosyncrasies which cannot be encoded in a context free setting of the suggestion grammar.

\noindent\textbf{Training data generation}
Our initial model was trained with a very constrained training set. We forced the training set to be limited to the minimal set of query patterns we wanted to support. In particular, queries followed a strict ordering of concepts. Instead of making the model more robust, it made it more brittle. We, since, switched to a training generation approach with more noise - randomly reordering terms and concepts while ensuring the data remains semantically meaningful in most cases.

Pre-trained embeddings were not sufficient in our context. The specificity of queries required embeddings trained for our domain.

The named entities that are normally used in NER systems were at times too specific for us: the difference between Person and Organization is less important because the ID Resolution module performs well even if these categories are merged.

\noindent\textbf{Project life cycle}
In an agile project setting, each component evolves and matures as the project progresses.
Some critical data science and engineering practices at our current stage would have been detrimental in the early pilot phases.
We now have a mature quality gating workflow with versioned GSD and AB experiment support, but it would have been unwise to spend effort on these early.

As adoption grows and users start to depend on the system, it is key to have those features and others like model versioning, metadata-driven algorithms, multi-stage quality pipeline, or extremely precise production instrumentation.

\section{Conclusions}

In this paper, we described an industrial NLS system integrated as part of Search of a major CRM platform. Our system was initially built with a deep learning NER model at its core. The architecture later evolved to better support the complexity and the real world constraints of enterprise customers.

Our NLS in its current state is a hybrid system that uses two tagging pipelines: one that combines Deep Learning with rule-based (expert) systems to generalize to various customers and, another that uses a grammar and mainly resolves our query suggestions. 
To use deep learning we overcame several challenges, the most important being, perhaps, the lack of training data. We found using synthetically generated, yet realistic, data generated by a PCFG to be very effective for our use-case.

To ensure high quality and  user satisfaction, before shipping a model we apply multiple validation checks. To track improvement in this constantly evolving setting we improved our model lifecycle management, our metadata and the evaluation datasets in ways we described in this paper. We now firmly believe that in the context of real applications investments in such tools are essential and should complement work on model improvements.

As adoption continues, we will add support for more concepts and new user intents, more customization (like custom objects that our customers can create) and more languages.
We will support more advanced remediations, and improve our system by learning specialized models for query suggestions, interpretations ranking and ID resolution.


\section{Acknowledgements}
We would like to thank Ahmet Bugdayci, Anmol Bhasin, Christian Posse, Dylan Hingey, Ghislain Brun, Mario Rodriguez, Paulo Gomes, Rohit Kapoor and Sam Edwards for their support, help and feedback throughout the development of this work.

\bibliographystyle{ACM-Reference-Format}
\bibliography{biblio}

\appendix

\end{document}